%% file: main.tex
\documentclass[10pt,twocolumn,letterpaper]{article}

\usepackage[pagenumbers]{wacv} 


\definecolor{wacvblue}{rgb}{0.21,0.49,0.74}
\usepackage[pagebackref,breaklinks,colorlinks,allcolors=wacvblue]{hyperref}

\usepackage{microtype}
\usepackage{multirow}
\usepackage[table]{xcolor}
\usepackage{makecell}
\usepackage{graphicx}
\usepackage{subcaption}
\usepackage{booktabs} 
\definecolor{goldenbrown}{RGB}{245, 222, 179}
\definecolor{lightgreen}{RGB}{220,240,220}
\usepackage{wasysym} 

\usepackage{pifont}
\newcommand{\cmark}{\ding{51}} 
\newcommand{\xmark}
{\ding{55}} 

\def\wacvPaperID{527} 
\def\confName{WACV}
\def\confYear{2027}

\title{AdaCount: Training-Free Similarity-Guided Spatial and Feature Adaptation for Zero-Shot Object Counting}

\author{
Muhammad Ibraheem Siddiqui \qquad
Muhammad Haris Khan \\
Mohamed Bin Zayed University of Artificial Intelligence \\
{\tt\small \{muhammad.siddiqui, muhammad.haris\}@mbzuai.ac.ae}
}

\begin{document}
\maketitle
\input{sec/0_abstract}    
\input{sec/1_intro}

\input{sec/2_Related_Work}

\input{sec/3_Method}

\input{sec/4_Experiments}

{
    \small
    \bibliographystyle{ieeenat_fullname}
    \bibliography{main}
}
\input{sec/5_Appendix}

\end{document}

%% file: sec/0_abstract.tex
\begin{abstract}
Zero-shot object counting (ZOC) aims to count instances of arbitrary object categories specified only through textual prompts. Recent training-free approaches leverage foundation models such as SAM to reformulate counting as a prompt-driven segmentation task, eliminating the need for costly counting-specific training data with point-level annotations. More recently, SAM3 introduced promptable concept segmentation, enabling the zero-shot segmentation of all instances corresponding to a text-defined concept. However, SAM3 struggles in densely populated scenes containing numerous small objects, where limited image resolution and insufficient attention to target-relevant regions often lead to missed instances and poor instance separation, hindering accurate object counting. To address this limitation, we propose AdaCount, a training-free framework for ZOC based on similarity-guided spatial and feature adaptation. AdaCount first estimates a prototype-driven similarity map that identifies target-relevant regions. This similarity map subsequently guides two complementary adaptations: (i) similarity-guided spatial warping, which reallocates image resolution toward target instances, and (ii) feature modulation, which amplifies target-relevant encoder representations. Together, these adaptations enable SAM3 to devote greater representational capacity to target-relevant regions while preserving global image context, without requiring any model retraining. Extensive experiments across six diverse counting benchmarks establish AdaCount as a new SOTA among training-free ZOC approaches. \href{https://muhammad-ibraheem-siddiqui.github.io/AdaCount/}{(Project-page)}
\end{abstract}

%% file: sec/1_intro.tex
\section{Introduction}
\label{sec:intro}

\begin{figure}[t]
    \centering
    \includegraphics[width=1\linewidth]{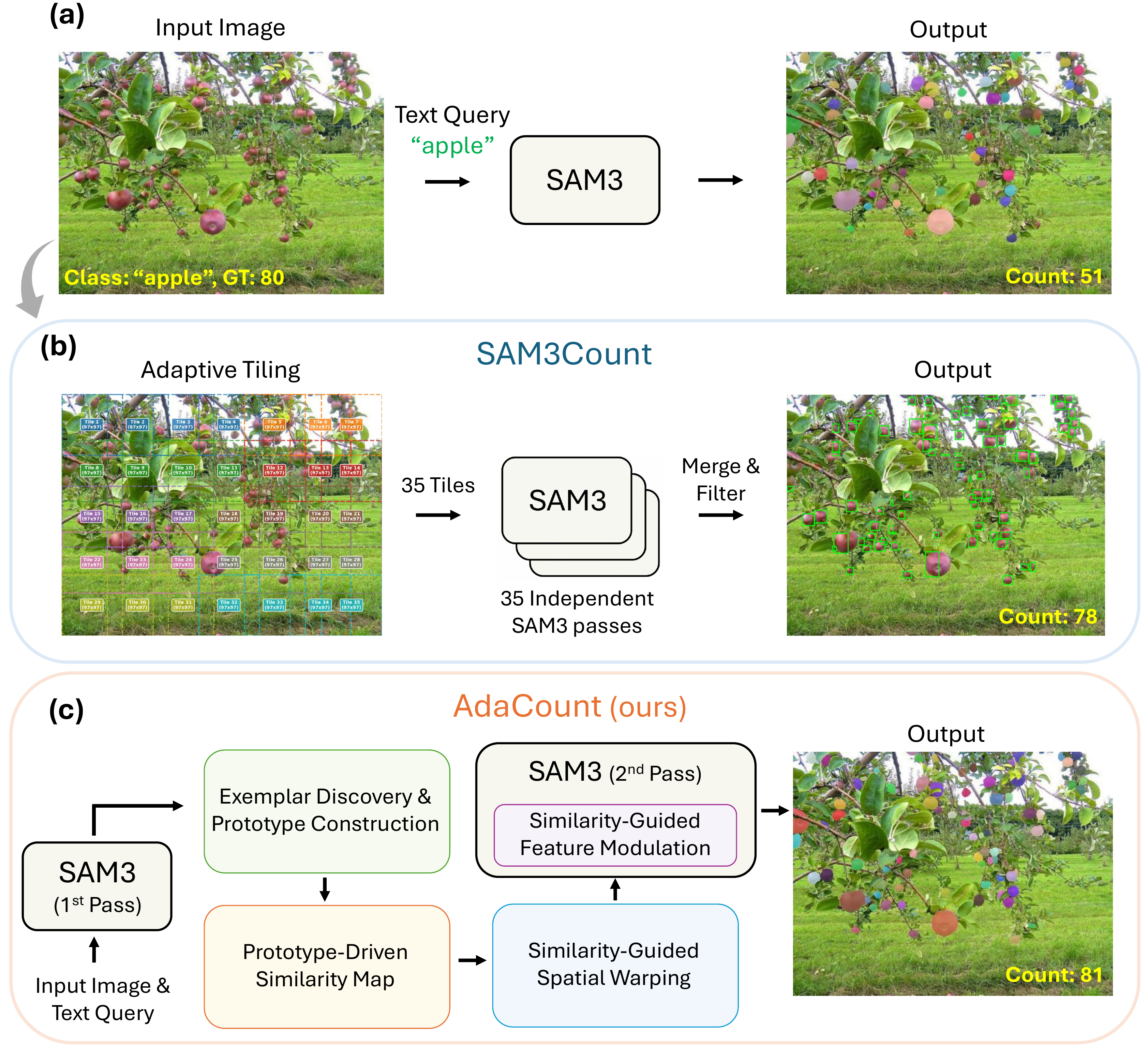}
    \caption{Comparison of SAM3, SAM3Count, and AdaCount (ours) for the ZOC task: (a) SAM3 substantially undercounts dense scenes, missing many target instances.
(b) SAM3Count addresses this limitation through adaptive tiling, requiring multiple independent SAM3 inference passes (35 in this case) over selected image regions. (c) In contrast, AdaCount introduces similarity-guided spatial and feature adaptation driven by a prototype-based similarity map. By reallocating representational capacity toward target-relevant regions, AdaCount substantially improves ZOC performance while requiring only two SAM3 inference passes.
}
    \label{fig:introfig}
\end{figure}

Object counting has attracted significant attention from the computer vision community due to its broad range of real-world applications~\cite{li2023calibrating, rahnemoonfar2017deep, bui2020vehicle}. In recent years, research has progressively shifted from class-specific counting methods~\cite{babu2022completely, li2020bi, tyagi2023degpr}, which rely on predefined object categories and require category-specific training, toward class-agnostic counting (CAC) approaches that aim to generalize to unseen object categories~\cite{ yang2021class,huang2023interactive,hui2024class}. Among CAC methods, few-shot counting leverages a small set of annotated exemplars to define the target category~\cite{nguyen2022few, you2023few, he2024few, yang2021class}. However, its dependence on exemplar annotations limits scalability and practical deployment~\cite{zhai2024zero, zhang2025enhancing}. In contrast, reference-less counting~\citep{hobley2022learning,d2024afreeca} eliminates the need for exemplars, but typically focuses on visually dominant patterns in a scene, making it difficult to selectively count instances of a user-specified category. To address these limitations, zero-shot object counting (ZOC) has emerged as a promising research direction~\cite{xu2023zero, zhu2024zero, jiang2023clip, qian2025t2icount,zhang2026boosting}. Given only a textual description of the target, ZOC aims to accurately count instances of arbitrary objects without requiring exemplar annotations.

Most existing ZOC approaches formulate counting as a density estimation problem~\cite{qian2025t2icount, xu2023zero, jiang2023clip, amini2023open, amini2024countgd}, learning a text-conditioned mapping from images to object counts using training data annotated with corresponding density maps. However, acquiring the point-level annotations required for density map generation is labor-intensive and difficult to scale across diverse object categories and domains, limiting the practical deployment of such methods in real-world applications. This limitation has motivated the emergence of training-free counting approaches~\cite{shi2024training, lin2025simple, mondal2025omnicount, ma2023can} that leverage the rich visual representations encoded within large pretrained foundation models such as SAM~\cite{kirillov2023segment}. Methods like TFOC~\cite{shi2024training} and OmniCount~\cite{mondal2025omnicount} reformulate ZOC as a text-driven segmentation task, where object counts are obtained by enumerating segmented instances. To compensate for the lack of semantic awareness in SAM~\cite{zhang2024personalize}, these methods introduce auxiliary semantic priors that guide class-specific mask generation and improve alignment with the target concept. More recently, SAM3~\cite{carion2025sam} has extended this paradigm by introducing a zero-shot promptable concept segmentation task, enabling the segmentation of all instances corresponding to a text-defined concept. Despite its strong zero-shot capabilities, SAM3 often struggles in densely populated scenes containing numerous small objects, Fig.~\ref{fig:introfig}(a). In such scenarios, limited image resolution and insufficient attention to target-relevant regions can lead to missed instances and poor instance separation, hindering accurate object counting. Concurrent work, SAM3Count~\cite{owusu2026sam3count}, improves SAM3 in dense scenes through an adaptive tiling strategy,  whereby selected image regions are processed independently by SAM3 at higher resolution, as illustrated in Fig.~\ref{fig:introfig}(b). While effective, this strategy relies on heuristic density-based criteria to determine when tile refinement should be applied and requires one SAM3 evaluation pass per tile, resulting in an inference cost that scales with the number of processed tiles.

To address the dense-scene limitations of SAM3, we propose AdaCount, a training-free framework for ZOC that leverages similarity-guided spatial and feature adaptation to improve counting performance, as illustrated in Fig.~\ref{fig:introfig}(c). At the core of AdaCount is a prototype-driven similarity map that identifies regions relevant to the target concept and guides adaptation in both the spatial and feature domains. In the spatial domain, we introduce similarity-guided spatial warping, which redistributes input image resolution such that target-relevant regions occupy a larger fraction of the warped image while less informative background regions are compressed. In the feature domain, we further modulate SAM3 encoder representations using the same prototype-driven similarity map through a residual gating strategy that amplifies feature responses within target-relevant regions. Together, these complementary adaptations enable SAM3 to devote greater representational capacity to target-relevant regions while preserving global contextual information. Extensive experiments across six diverse counting benchmarks demonstrate that AdaCount achieves SOTA performance among training-free ZOC approaches. In summary:

\begin{itemize}

\item We introduce AdaCount, a training-free framework that improves zero-shot object counting by leveraging SAM3 together with a prototype-driven similarity map to guide adaptation in the spatial and feature domains.

\item We propose the following similarity-guided adaptations: (i) a spatial warping mechanism that reallocates image resolution toward target-relevant regions, and (ii) a feature-level modulation strategy that amplifies target-relevant encoder representations.

\item We conduct extensive experiments across six diverse counting benchmarks, where AdaCount consistently achieves superior ZOC performance over existing approaches.

\end{itemize}

%% file: sec/2_Related_Work.tex
\section{Related Work}
\label{sec:related_work}


\noindent\textbf{Zero-Shot Object Counting.} Existing ZOC approaches can be broadly categorized into two directions. The first focuses on learning image-text alignment to understand object-related correlations without requiring physical exemplars~\cite{kang2024vlcounter, amini2023open, jiang2023clip, qian2025t2icount, zhang2026boosting}. The second seeks to automatically discover class-relevant exemplars given only a textual prompt~\cite{xu2023zero,zhu2024zero, liu2025countse}. Despite their class-agnostic nature, these approaches typically require training on counting datasets containing point-level annotations~\cite{ranjan2021learning}. Acquiring such annotations is labor-intensive for large-scale datasets containing densely populated scenes, motivating the development of training-free alternatives.

\noindent\textbf{Training-Free Zero-Shot Object Counting.} Training-free approaches to ZOC exploit the rich visual representations encoded within large pretrained foundation models~\cite{shi2024training}. In particular, the introduction of SAM~\cite{kirillov2023segment} transformed ZOC into a prompt-driven segmentation task, where object counts are obtained by enumerating segmented instances. Methods such as TFOC~\cite{shi2024training} augment SAM with semantic priors to compensate for its class-agnostic nature and improve alignment between predicted masks and the target concept. OmniCount~\cite{mondal2025omnicount} further extends this paradigm by incorporating geometric priors alongside semantic cues to improve counting performance. More recently, SAM3~\cite{carion2025sam} introduced a promptable concept segmentation task that explicitly enables the zero-shot segmentation of all instances corresponding to a text-defined concept, mitigating the need for auxiliary semantic guidance in SAM. Despite its strong zero-shot capabilities, SAM3 often struggles in dense scenes containing numerous small objects. Concurrent work, SAM3Count~\cite{owusu2026sam3count}, addresses this limitation through adaptive tiling strategy, whereby selected image regions are independently processed by SAM3 at higher resolution. While effective, the number of SAM3 evaluations scales with the number of generated tiles and can further increase in highly crowded scenes where recursive refinement is required, leading to prohibitive computational overhead. Moreover, the approach relies on a heuristic tiling criterion to determine where refinement should be applied. In contrast, AdaCount employs a fixed two-pass pipeline that leverages a prototype-driven similarity map to guide spatial and feature adaptation, enabling improved counting performance while maintaining a predictable inference cost independent of scene density.

\noindent\textbf{Content-Adaptive Pixel-Level Image Warping.}
Content-adaptive image warping reallocates image resolution toward visually important regions. Early approaches include seam carving~\cite{rubinstein2010comparative}, saliency-aware warping~\cite{yoo2013content}, mesh-based image retargeting~\cite{guo2009image}, and energy-minimization techniques~\cite{karni2009energy}, while more recent methods have explored adaptive resizing~\cite{talebi2021learning}, domain adaptation~\cite{zheng2025instance}, and perceptual magnification~\cite{mao2025through}. Most recently, AttWarp~\cite{dalal2025constructive} introduced inference-time attention-guided warping for multimodal large language models (MLLMs), where image resolution is redistributed according to query-conditioned attention maps. However, MLLM attention maps are often dominated by a few highly attended regions~\cite{bi2025unveiling}, causing resolution to be concentrated on a small subset of target locations. While suitable for tasks where identifying a few query-relevant regions is sufficient to answer the question, this behavior is less desirable for dense counting, where potentially many target instances distributed throughout the scene must be simultaneously preserved. To address this limitation, AdaCount derives the warping signal from a prototype-driven similarity map rather than decoder attention (Fig.~\ref{fig:attvssim}). Since similarity is evaluated independently at every spatial location, multiple target instances can receive strong responses concurrently, producing a more distributed importance field that reallocates resolution across all candidate instances.

%% file: sec/3_Method.tex
\section{Method}
\label{sec:method}

\begin{figure*}[t]
    \centering
    \includegraphics[width=0.97\linewidth]{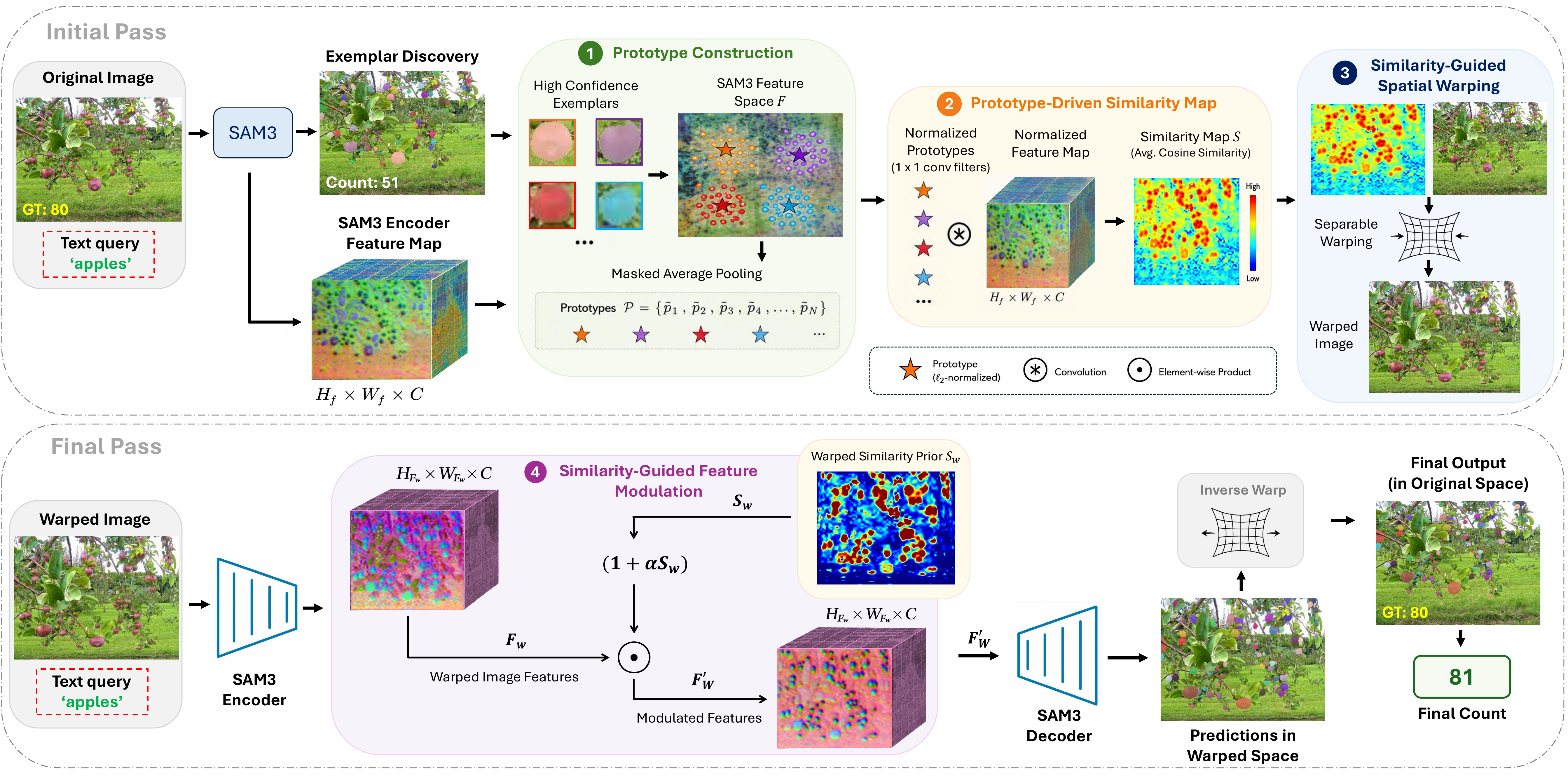}
    \caption{Overview of AdaCount: Given an input image and text query, AdaCount first performs an initial SAM3 pass to discover high-confidence target exemplars. These exemplars are used to construct image-specific prototypes in the SAM3 feature space via masked average pooling. The prototypes are then correlated with encoder features to produce a similarity map that identifies target-relevant regions. The resulting map guides spatial warping of the input image before a second SAM3 inference pass. The warped image is then processed by the SAM3 encoder, whose features are subsequently modulated using the warped similarity prior before being forwarded to the decoder. Finally, predictions obtained in the warped space are mapped back to the original space to produce the final output. }
    \label{fig:mainfig}
\end{figure*}


\subsection{Exemplar Discovery \& Prototype Construction}

Fig.~\ref{fig:mainfig} illustrates the overview of AdaCount. Given an input image $I \in \mathbb{R}^{H\times W\times 3}$ and a text query $q$ describing the target object category, AdaCount first performs an initial object discovery pass using SAM3. This pass provides a reliable set of high-confidence target instances, which are subsequently used as exemplars to construct prototype representations. Specifically, SAM3 produces an initial set $\mathcal{D}$ of high-confidence instance predictions given as:

\begin{equation}
\mathcal{D}
=
\left\{
(M_i,B_i,s_i)
\right\}_{i=1}^{N},
\end{equation} where $M_i$, $B_i$, and $s_i$ denote the predicted mask, bounding box, and confidence score of the $i$-th instance, respectively. Following the SAM3 post-processing step, duplicate and overlapping predictions are suppressed using non-maximal suppression based on Intersection-over-Minimum (IoM). These predictions serve as exemplars that provide image-specific evidence of the target category and form the basis for subsequent prototype construction. Next, we extract the encoder feature map from SAM3,

\begin{equation}
\mathbf{F}
=
f_{\text{enc}}(I)
\in
\mathbb{R}^{C\times H_f\times W_f},
\end{equation} where $f_{\text{enc}}(\cdot)$ denotes the SAM3 image encoder (which essentially is an aligned vision-language perception encoder~\cite{bolya2026perception}), $C$ is the feature dimension, and $H_f \times W_f$ is the spatial resolution of the feature map. For each exemplar, the corresponding segmentation mask is mapped to the encoder feature resolution and used to identify the feature vectors belonging to the target instance. These feature vectors are subsequently aggregated through masked average pooling to construct a prototype representation $\mathbf{p}_i \in \mathbb{R}^{C}$.

\begin{equation}
\mathbf{p}_i
=
\frac{
\sum\limits_{(u,v)\in M_i}
\mathbf{F}(u,v)
}
{
|M_i|
},
\end{equation} where $|M_i|$ is the number of feature locations associated with exemplar mask. Each prototype is subsequently $\ell_2$-normalized, resulting in set $\mathcal{P}$ of image-specific prototypes

\begin{equation}
\tilde{\mathbf{p}}_i
=
\frac{\mathbf{p}_i}
{\|\mathbf{p}_i\|_2+\epsilon},
\qquad
\mathcal{P}
=
\left\{
\tilde{\mathbf{p}}_i
\right\}_{i=1}^{N}.
\end{equation}

\subsection{Prototype-Driven Similarity Map}

The prototype set $\mathcal{P}$ is subsequently used to identify target relevant regions. To this end, we compute the similarity between the prototypes and the SAM3 feature map. First, the encoder feature vectors are $\ell_2$-normalized,

\begin{equation}
\tilde{\mathbf{F}}(u,v)
=
\frac{\mathbf{F}(u,v)}
{\|\mathbf{F}(u,v)\|_2+\epsilon},
\end{equation} where $\tilde{\mathbf{F}}(u,v)\in\mathbb{R}^{C}$ denotes the normalized feature vector at location $(u,v)$. Given a normalized prototype $\tilde{\mathbf{p}}_i$, the corresponding similarity response is computed by measuring the cosine similarity between the prototype and every spatial location in the feature map:

\begin{equation}
S_i(u,v)
=
\tilde{\mathbf{p}}_i^{\top}
\tilde{\mathbf{F}}(u,v),
\end{equation} where $S_i(u,v)$ maps the similarity between the $i$-th prototype and the feature vector at location $(u,v)$. Since both the prototype and feature vectors are $\ell_2$-normalized, the inner product is equivalent to cosine similarity. In practice, the cosine similarity responses are computed in an efficient manner by treating the normalized prototypes as a bank of $1\times1$ convolutional filters and correlating them with the normalized feature map in a single operation. The resulting prototype-specific responses are subsequently aggregated to obtain a unified similarity representation $S$:

\begin{equation}
S(u,v)
=
\frac{1}{N}
\sum_{i=1}^{N}
S_i(u,v),
\end{equation} where $N$ denotes the number of prototypes. The aggregated similarity map highlights target-relevant regions and serves as the basis for the subsequent similarity-guided adaptations.

\subsection{Similarity-Guided Adaptations}

We leverage the similarity map $S$ to adapt the input processed by SAM3. Specifically, the similarity prior is first used to construct a non-uniform spatial transformation that reallocates input image resolution toward target-relevant regions. The resulting warped representation is then further refined through feature modulation guided by the same similarity prior. Together, these two complementary mechanisms enable AdaCount to devote greater representational capacity to target-relevant regions while preserving the global context.

\begin{figure}[t]
    \centering
    \includegraphics[width=1\linewidth]{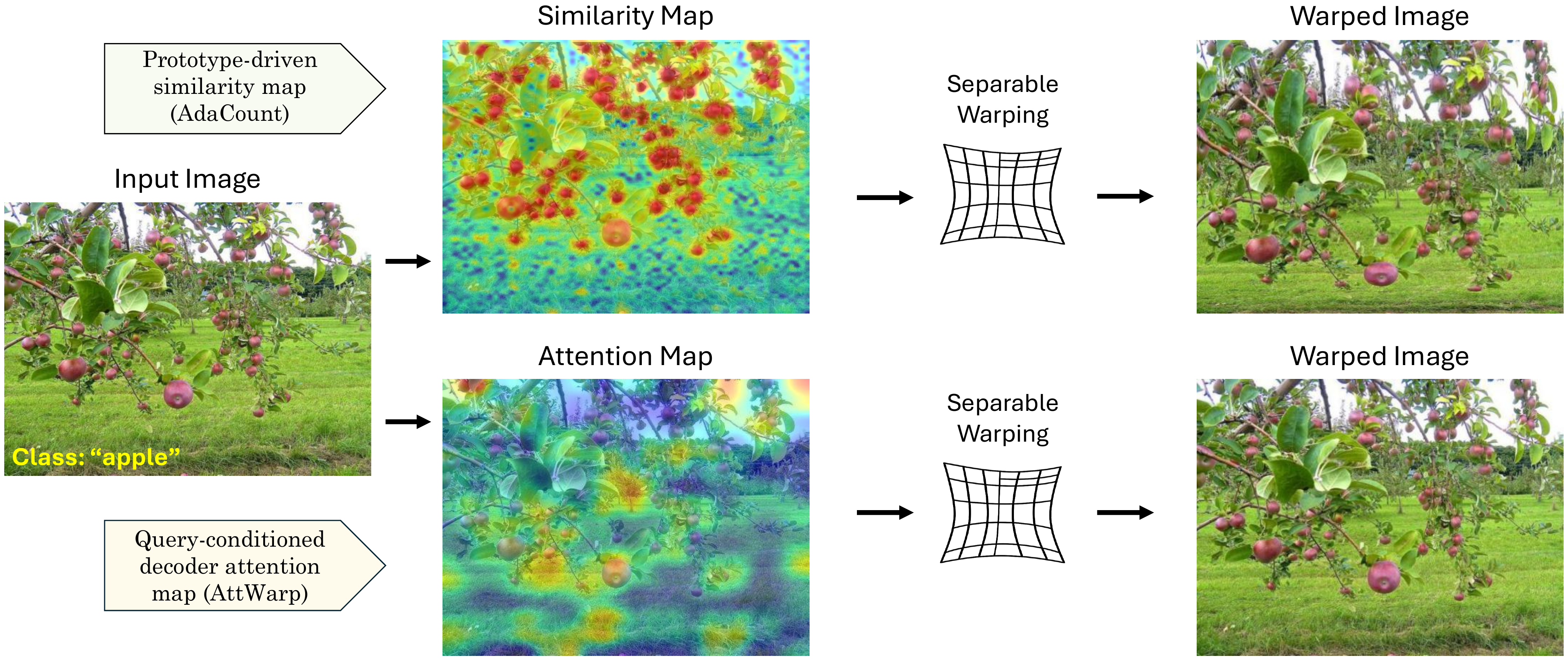}
    \caption{
Attention-guided versus prototype-driven similarity-guided warping. AttWarp~\cite{dalal2025constructive} derives its rectilinear warp from query-conditioned attention maps extracted from a MLLM (LLaVA~\cite{liu2024improved} in this example), producing a relatively diffuse importance field focused on a few dominant regions. In contrast, AdaCount constructs a prototype-driven similarity map from SAM3-discovered exemplars yielding localized, instance-aware responses. Consequently, AdaCount reallocates spatial resolution across all candidate objects rather than a small set of highly attended regions, making it particularly effective for dense object counting.
}
    \label{fig:attvssim}
\end{figure}

\noindent\textbf{Similarity-Guided Spatial Warping.} Our warping formulation adopts a rectilinear warping strategy inspired by AttWarp~\cite{dalal2025constructive}, while deriving the warping signal from prototype-driven similarity rather than query-conditioned decoder attention (Fig.~\ref{fig:attvssim}).
Given the similarity map $S$, we first sharpen and smooth the similarity responses. Specifically, a power transformation followed by Gaussian filtering is applied to increase the contrast of the similarity map by suppressing weak and moderate similarity responses relative to highly confident regions, i.e., $\bar{S}=G_{\sigma}*S^{\gamma}$, where $\gamma$ controls the sharpening strength and $G_{\sigma}$ denotes a Gaussian kernel.  The similarity map determines how much spatial area each image location should occupy after warping. Larger values indicate regions that should be expanded, while smaller values indicate regions that may be compressed. Let $x \in \{1,\ldots,W\}$ and $y \in \{1,\ldots,H\}$ denote horizontal and vertical image coordinates, respectively. To construct an efficient separable warp, we first compute the horizontal and vertical marginal importance distributions by accumulating the similarity values along each axis:\begin{equation}
m_x(x)
=
\sum_{y=1}^{H} \bar{S}(y,x),
\qquad
m_y(y)
=
\sum_{x=1}^{W} \bar{S}(y,x),
\end{equation} where $m_x$ and $m_y$ represent the total importance mass for each row and column, respectively. Next, we convert these into normalized cumulative distribution functions (CDFs):
\begin{equation}
C_x(x)
=
\frac{\sum_{t=1}^{x} m_x(t)}
{\sum_{t=1}^{W} m_x(t)},
\quad
C_y(y)
=
\frac{\sum_{t=1}^{y} m_y(t)}
{\sum_{t=1}^{H} m_y(t)}.
\end{equation} The resulting CDFs define a separable warp grid with the same spatial resolution as the original image. Since the marginal importance distributions are strictly positive, the resulting CDFs are monotonic and therefore invertible. This property enables predictions obtained in the warped domain to be mapped back to the original image coordinates during the final inference stage. Rather than increasing the image size, the proposed transformation redistributes the available pixel budget within a fixed-resolution image canvas, allocating more samples to regions with higher importance and fewer samples to less relevant regions.

The inverse CDFs determine the source coordinates associated with each location in the warped image. For a uniformly sampled target coordinate $(u,v)$ in the warped domain, the corresponding source coordinate is obtained as

\begin{equation}
x_s
=
C_x^{-1}(u),
\qquad
y_s
=
C_y^{-1}(v),
\label{eq:inverse}
\end{equation} where $(x_s,y_s)$ denotes the corresponding sampling location in the original image. This defines the similarity-guided warping operator $\mathcal{W}$ as:

\begin{equation}
I_w(u,v)
=
\mathcal{W}(I,\bar{S})(u,v)
=
I(x_s,y_s),
\label{eq:inverse_operator}
\end{equation} implemented using bilinear interpolation. As a result, regions exhibiting strong $S$ responses occupy a larger portion of the warped image while preserving the global context.

\noindent\textbf{Sim-Guided Feature Modulation \& Final Inference.}
The warped image $I_w$ serves as the input to the second SAM3 inference pass $\mathbf{F}_w
=f_{\text{enc}}(I_w)$, where $\mathbf{F}_w$ represents the encoder feature map. To further exploit the similarity map $S$, AdaCount performs feature modulation directly on $\mathbf{F}_w$ before it is forwarded to the decoder. To ensure consistency, $S$ is first transformed into the warped space using the same warping operator $\mathcal{W}$. This produces a warped similarity prior $S_w$ that is spatially aligned with the warped image $(I_w)$.  Rather than using $S_w$ as a hard attention mask, we employ a residual gating strategy that preserves the original feature information while selectively emphasizing target-relevant regions. Specifically, $S_w$ is centered around its spatial mean,

\begin{equation}
\hat{S}_w
=
S_w
-
\operatorname{mean}(S_w),
\end{equation} and subsequently injected into the encoder features through

\begin{equation}
\mathbf{F}'_w
=
\mathbf{F}_w
\odot
\left(
1+\alpha \hat{S}_w
\right),
\end{equation} where $\odot$ denotes element-wise product and $\alpha$ controls the modulation strength. The modulated feature representation $\mathbf{F}'_w$ is then forwarded to the SAM3 decoder for mask prediction in the warped space. This formulation amplifies the magnitude of feature responses in regions exhibiting strong similarity to the discovered exemplars while preserving the original SAM3 features, thereby avoiding the information loss associated with hard masking or feature replacement. Consequently, AdaCount combines two complementary forms of adaptation: similarity-guided spatial warping reallocates image samples toward target regions, while feature modulation enhances representational prominence within the SAM3 feature space. The final SAM3 predictions in the warped space are subsequently mapped back to the original space using the inverse coordinate mapping defined in Eq.~\ref{eq:inverse}.

%% file: sec/4_Experiments.tex
\section{Experiments}
\label{sec:experiments}

\subsection{Setup}

\noindent\textbf{Datasets.} We evaluate AdaCount on six diverse counting benchmarks: (1) FSC-147~\cite{ranjan2021learning} contains 6,135 images covering 147 object categories, with object counts ranging from 7 to 3,731 instances per image and an average of 56. (2) CARPK~\cite{hsieh2017drone} consists of 1,448 drone-view parking-lot images containing 89,777 cars, with 459 images reserved for testing. (3) PerSense-D~\cite{siddiqui2025persense} comprises 717 images spanning 28 object categories, with an average of 53 target instances per image. (4) OmniCount-191~\cite{mondal2025omnicount} is a multi-class counting benchmark.  Following prior work, we evaluate on its fruit subset, which consists of 303 images containing 8 different fruit categories. (5) MBM (modified bone marrow) dataset~\cite{paul2017count} contains 44 microscopy images with (126 ± 33) nuclei per image, presenting a challenging counting scenario due to substantial appearance variability. (6) PrACo~\cite{ciampi2025mind} is a prompt-aware counting benchmark built upon FSC-147 that evaluates a model’s ability to correctly interpret textual prompts. It introduces a Negative-Label Test, where images are queried with object categories absent from the scene, and a Mosaic Test, where multi-class images are artificially created by stitching together images from different object categories, requiring the model to count only the category specified by the prompt while ignoring distractor objects.

\noindent\textbf{Evaluation Metrics.} Following prior ZOC works, we report the mean absolute error (MAE) and root mean square error (RMSE) to evaluate counting accuracy and robustness, respectively. For the PrACo benchmark, we additionally report the Normalized Mean of Negative Predictions (NMN), Positive Class Count Nearness (PCCN), Counting Precision (CntPr), Counting Recall (CntR), and overall Counting F1 score (CntF1) metrics proposed in~\cite{ciampi2025mind}.

\noindent\textbf{Implementation Details.} AdaCount is built upon the publicly available SAM3 implementation, requiring no additional training or fine-tuning. Input images are resized to the default SAM3 resolution of 1008 $\times$ 1008, and textual prompts are derived directly from the category names provided by each dataset. The similarity sharpening parameter ($\gamma$), feature modulation strength ($\alpha$), and Gaussian smoothing kernel ($G_{\sigma}$) are fixed at 2, 0.9, and 11 $\times$ 11, respectively, based on the ablations in Sec.~\ref{sec:ablations}. All hyperparameters are fixed across all datasets with no dataset-specific tuning.


\begin{table}[t]
\centering
\setlength{\tabcolsep}{1.8pt}
\fontsize{8pt}{8pt}\selectfont
\caption{Results on the FSC-147 dataset. Best and second-best results are highlighted in bold and underlined, respectively. * denotes concurrent work. $^{\dagger}$ indicates a few-shot counting method requiring visual exemplars at inference time.}
\begin{tabular}{llccc}
\toprule
\rowcolor{lightgray!30} 
Method & Venue & Training & MAE~$\downarrow$ & RMSE~$\downarrow$ \\
\midrule
ZSOC~\cite{xu2023zero} & CVPR'23 & \cmark & 22.09 & 115.17 \\
CLIP-Count~\cite{jiang2023clip} & ACM'23 & \cmark & 17.78 & 106.62 \\
CounTX~\cite{amini2023open} & BMVC'23 & \cmark & 15.88 & 106.29 \\
VLCounter~\cite{kang2024vlcounter} & AAAI'24 & \cmark & 17.05 & 106.16 \\
PseCo~\cite{huang2024point} & CVPR'24 & \cmark & 16.58 & 129.77 \\
DAVE~\cite{pelhan2024dave} & ICCV'24 & \cmark & 14.90 & 103.42 \\
VA-Count~\cite{zhu2024zero} & ECCV'24 & \cmark & 17.88 & 129.31 \\
CountGD~\cite{amini2024countgd} & NeurIPS'24 & \cmark & 14.76 & 120.42 \\
GeCo~\cite{pelhan2024novel} & NeurIPS'24 & \cmark & 13.30 & 108.72 \\
T2ICount~\cite{qian2025t2icount} & CVPR'25 & \cmark & \textbf{11.76} & 97.86 \\
Yolo-Count~\cite{zeng2025yolo} & ICCV'25 & \cmark & 14.80 & \textbf{96.14} \\
QICA~\cite{zhang2026boosting} & CVPR'26 & \cmark & \underline{12.41} & \underline{97.28} \\
CountGD++~\cite{amini2026countgd++} & CVPR'26 & \cmark & 16.55 & 129.76 \\

\midrule
CountingDINO$^\dagger$~\cite{pacini2025countingdino} & WACV'26 & \xmark & 20.93 & 71.37 \\
\midrule
SAM~\cite{kirillov2023segment} & ICCV'23 & \xmark & 42.48 & 137.50 \\
Count Anything~\cite{ma2023can} & arXiv'23 & \xmark & 27.97 & 131.24 \\
G-DINO~\cite{liu2024grounding} & ECCV'24 & \xmark & 59.23 & 159.28 \\
TFOC~\cite{shi2024training} & WACV'24 & \xmark & 24.79 & 137.15 \\
OmniCount~\cite{mondal2025omnicount} & AAAI'25 & \xmark & 21.46 & 133.28 \\
SAM3~\cite{carion2025sam} & ICLR'26 & \xmark & 22.89 & 144.80 \\
SAM3Count*~\cite{owusu2026sam3count} & CVPRW'26 & \xmark & \underline{19.80} & \underline{128.59} \\
\rowcolor{lightgreen!50}
AdaCount & (Ours) & \xmark & \textbf{18.97} & \textbf{126.94} \\
\bottomrule
\end{tabular}
\label{tab:fsc147_eval}
\end{table}

\begin{table}[t]
\centering
\caption{Results on [Left]: CARPK, [Right]: OmniCount (Fruits) }
\label{tab:combined}
\footnotesize

\begin{subtable}[t]{0.5\columnwidth}
\centering
\setlength{\tabcolsep}{1pt}
\label{tab:carpk_eval}

\resizebox{\linewidth}{!}{%
\begin{tabular}{lcc}
\toprule
\rowcolor{lightgray!30}
Method & MAE~$\downarrow$ & RMSE~$\downarrow$ \\
\midrule
VLCounter~\cite{kang2024vlcounter} & 6.46 & 8.68 \\
GeCo~\cite{pelhan2024novel} & 10.34 & 14.73 \\
CountGD~\cite{amini2024countgd} & \underline{3.83} & \underline{5.41} \\
T2ICount~\cite{qian2025t2icount} & 8.61 & 13.47 \\
QICA~\cite{zhang2026boosting} & 6.07 & 7.82 \\
CountGD++~\cite{amini2026countgd++} & \textbf{2.48} & \textbf{3.74} \\
\midrule
TFOC~\cite{shi2024training} & 14.35 & 17.22 \\
OmniCount~\cite{mondal2025omnicount} & 13.41 & 16.85 \\
CountingDINO~\cite{pacini2025countingdino} & 21.26 & 28.20 \\
SAM3~\cite{carion2025sam} & 3.52 & 6.57 \\
SAM3Count~\cite{owusu2026sam3count} & \underline{3.11} & \underline{5.60} \\
\rowcolor{lightgreen!50}
AdaCount (Ours) & \textbf{1.99} & \textbf{3.05} \\
\bottomrule
\end{tabular}
}
\end{subtable}
\hfill
\begin{subtable}[t]{0.47\columnwidth}
\centering
\setlength{\tabcolsep}{1pt}
\label{tab:omnicount_fruits}

\resizebox{\linewidth}{!}{%
\begin{tabular}{lcc}
\toprule
\rowcolor{lightgray!30}
Method & MAE~$\downarrow$ & RMSE~$\downarrow$ \\
\midrule
CountGD~\cite{amini2024countgd} & 2.76 & 3.11 \\
CountGD-Box~\cite{amini2026open} & 2.83 & 3.15 \\
CountGD++~\cite{amini2026countgd++} & \textbf{1.97} & \textbf{3.29} \\
\midrule
SAM3~\cite{carion2025sam} & 0.44 & 0.93 \\
SAM3Count~\cite{owusu2026sam3count} & 0.43 & 0.93 \\
\rowcolor{lightgreen!50}
AdaCount (Ours) & \textbf{0.37} & \textbf{0.74} \\
\bottomrule
\end{tabular}
}
\vspace{2mm}

\parbox{\linewidth}{\raggedright
\scriptsize
Note: In both tables, methods above the horizontal separator are training-based, whereas methods below the separator are training-free.}
\end{subtable}

\end{table}

\begin{table}[t]
\centering
\setlength{\tabcolsep}{2pt}
\fontsize{8pt}{8pt}\selectfont
\caption{Results on MBM and PerSense-D datasets.}
\begin{tabular}{lcc|cc|cc}
\toprule
\rowcolor{lightgray!30}
& & \multicolumn{2}{c}{MBM} & \multicolumn{2}{c}{PerSense-D} \\
\cmidrule(lr){3-4} \cmidrule(lr){5-6}
Method & Training & MAE~$\downarrow$ & RMSE~$\downarrow$ & MAE~$\downarrow$ & RMSE~$\downarrow$ \\
\midrule
GeCo~\cite{pelhan2024novel} & \cmark & 92.59 & 104.52 & 13.12 & 28.36  \\
T2ICount~\cite{qian2025t2icount} & \cmark & 30.01 & 44.44 & 24.95 & 88.21 \\

\midrule
TFOC~\cite{shi2024training} & \xmark & 64.25 & 75.67 & 17.79 & 35.74 \\
SAM3~\cite{carion2025sam} & \xmark & 23.25 & 32.08 & 10.66 & 31.14 \\
SAM3Count~\cite{owusu2026sam3count} & \xmark & 56.13 & 59.35 & 8.97 & 26.76 \\
\rowcolor{lightgreen!50}
AdaCount (Ours) & \xmark & \textbf{19.11} & \textbf{21.69} & \textbf{7.51} & \textbf{25.35} \\
\bottomrule
\end{tabular}
\label{tab:mbm_persensed}
\end{table}

\begin{table}[t]
\centering
\setlength{\tabcolsep}{3pt}
\fontsize{8pt}{8pt}\selectfont
\caption{Results on the PrACo benchmark.}
\begin{tabular}{lcccccc}
\toprule
\rowcolor{lightgray!30}
& &
\multicolumn{2}{c}{Negative Test} &
\multicolumn{3}{c}{Mosaic Test} \\
\cmidrule(lr){3-4} \cmidrule(lr){5-7}
Method & Training & NMN$\downarrow$ & PCCN$\uparrow$ & CntP$\uparrow$ & CntR$\uparrow$ & CntF1$\uparrow$ \\
\midrule
CounTX~\cite{amini2023open} & \cmark & 0.95 & \textbf{64.51} & 0.68 & 0.71 & 0.69 \\
CLIP-Count~\cite{jiang2023clip} & \cmark & 1.27 & 38.13 & 0.49 & 0.76 & 0.59 \\
VLCounter~\cite{kang2024vlcounter} & \cmark & 1.15 & 53.36 & 0.51 & 0.78 & 0.62 \\
DAVE~\cite{pelhan2024dave} & \cmark & 1.05 & 37.02 & 0.84 & 0.80 & 0.81 \\
CountGD++~\cite{amini2026countgd++} & \cmark & \textbf{0.88} & 62.86 & \textbf{0.86} & \textbf{0.96} & \textbf{0.90} \\
\midrule
TFOC & \xmark & 0.75 & 66.04 & 0.68 & 0.84 & 0.75 \\
SAM3 & \xmark & \textbf{0.01} & 93.70 & 0.79 & 0.93 & 0.85 \\
\rowcolor{lightgreen!50}
AdaCount (Ours) & \xmark & 0.02 & \textbf{95.21} & \textbf{0.84} & \textbf{0.95} & \textbf{0.89} \\
\bottomrule
\end{tabular}
\label{tab:praco_results}
\end{table}

\begin{table*}[t]
\centering
\footnotesize

\caption{Ablation studies of AdaCount on the FSC-147 benchmark. (a) Contribution of the proposed adaptations. (b--d) Sensitivity to similarity sharpening parameter \(\gamma\), feature modulation strength \(\alpha\), and Gaussian smoothing kernel \(G_{\sigma}\), respectively.}

\vspace{-2mm}

\begin{subtable}[t]{0.29\textwidth}
\fontsize{9pt}{9pt}\selectfont
\centering
\caption{}
\label{tab:adacount_components}
\setlength{\tabcolsep}{1.8pt}
\resizebox{\linewidth}{!}{%
\begin{tabular}{lcccc}
\toprule
\rowcolor{lightgray!30}
Method &
\begin{tabular}{c}Spatial\\Warping\end{tabular} &
\begin{tabular}{c}Feature\\Modulation\end{tabular} &
MAE &
RMSE \\
\midrule
SAM3 & \xmark & \xmark & 22.89 & 144.80 \\
AdaCount & \cmark & \xmark & 19.90 & 129.10 \\
\rowcolor{lightgreen!50}
AdaCount & \cmark & \cmark & 18.97 & 126.94 \\
\bottomrule
\end{tabular}}
\end{subtable}
\hfill
\begin{subtable}[t]{0.18\textwidth}
\fontsize{9pt}{9pt}\selectfont
\centering
\caption{}
\label{tab:gamma_analysis}
\setlength{\tabcolsep}{1.8pt}
\resizebox{\linewidth}{!}{%
\begin{tabular}{cccc}
\toprule
\rowcolor{lightgray!30}
$\gamma$ & 1 & 2 & 3 \\
\midrule
MAE & 19.15 & \cellcolor{lightgreen!50}18.97 & 19.33 \\
RMSE & 127.33 & \cellcolor{lightgreen!50}126.94 & 128.71 \\
\bottomrule
\end{tabular}}
\end{subtable}
\hfill
\begin{subtable}[t]{0.26\textwidth}
\fontsize{9pt}{9pt}\selectfont
\centering
\caption{}
\label{tab:alpha_analysis}
\setlength{\tabcolsep}{1.8pt}
\resizebox{\linewidth}{!}{%
\begin{tabular}{cccccc}
\toprule
\rowcolor{lightgray!30}
$\alpha$ & 0.25 & 0.50 & 0.75 & 0.90 & 0.95 \\
\midrule
MAE & 19.56 & 19.28 & 19.02 & \cellcolor{lightgreen!50}18.97& 19.00 \\
RMSE & 128.65 & 127.72 & 127.40 & \cellcolor{lightgreen!50}126.94 & 127.01 \\
\bottomrule
\end{tabular}}
\end{subtable}
\hfill
\begin{subtable}[t]{0.23\textwidth}
\fontsize{9pt}{9pt}\selectfont
\centering
\caption{}
\label{tab:gaussian_analysis}
\setlength{\tabcolsep}{1.8pt}
\resizebox{\linewidth}{!}{%
\begin{tabular}{ccccc}
\toprule
\rowcolor{lightgray!30}
$G_\sigma$ & 3$\times$3 & 7$\times$7 & 11$\times$11 & 13$\times$13 \\
\midrule
MAE & 19.29 & 19.15 & \cellcolor{lightgreen!50}18.97 & 19.01 \\
RMSE & 127.30 & 127.18 & \cellcolor{lightgreen!50}126.94 & 127.21 \\
\bottomrule
\end{tabular}}
\end{subtable}
\vspace{-2mm}

\label{tab:ablation_studies}
\end{table*}

\subsection{Results}
\label{sec:results}

\noindent\textbf{FSC-147.} Table~\ref{tab:fsc147_eval} compares AdaCount with state-of-the-art (SOTA) ZOC methods on the FSC-147 benchmark. AdaCount substantially improves the zero-shot counting performance compared to SAM3, reducing MAE and RMSE by 17.12\% and 12.33\%, respectively. Furthermore, AdaCount achieves SOTA performance among existing training-free ZOC approaches, demonstrating the effectiveness of the proposed similarity-guided spatial and feature adaptations.

\noindent\textbf{CARPK.} We evaluate AdaCount on the CARPK dataset using \emph{``cars''} as the textual prompt. As shown in Table~\ref{tab:combined} (left), AdaCount substantially improves upon the SAM3 baseline, reducing MAE and RMSE by 43.46\% and 53.57\%, respectively. Notably, AdaCount also outperforms QICA~\cite{zhang2026boosting} and CountGD++~\cite{amini2026countgd++}, both of which are current SOTA among the training-based approaches. These results highlight the superiority of our AdaCount, establishing it as the best-performing method among all compared approaches on CARPK dataset.

\noindent\textbf{OmniCount (Fruits).} Following the evaluation protocol of CountGD++~\cite{amini2026countgd++}, we evaluate AdaCount on the fruit subset of the OmniCount benchmark. As illustrated in Table~\ref{tab:combined} (right), AdaCount improves upon the SAM3 baseline, reducing MAE and RMSE by 15.90\% and 20.43\%, respectively. In contrast, the concurrent SAM3Count method performs comparable to SAM3, as its tile refinement strategy is primarily activated only in dense scenes and therefore provides limited benefit on the relatively sparse OmniCount (fruits) subset. These results demonstrate that AdaCount is effective not only in dense scenes but also in multi-object sparse scenarios. Notably, AdaCount establishes a new SOTA on this benchmark despite relying solely on textual prompts, whereas competing training-based approaches additionally require visual exemplar prompts at inference time.

\noindent\textbf{MBM and PerSense-D.} As summarized in Table~\ref{tab:mbm_persensed}, AdaCount on MBM benchmark reduces MAE and RMSE by 17.80\% and 32.38\%, respectively, compared to the SAM3 baseline, outperforming all competing methods. Interestingly, concurrent work SAM3Count degrades performance relative to SAM3 on this dataset. Careful investigation attributes this behavior to its tile-based refinement strategy, where small image regions are independently resized and processed at the default SAM3 resolution. In microscopy images like MBM, such aggressive upsampling can distort fine cellular structures and remove important contextual information, adversely affecting segmentation quality. In contrast, AdaCount preserves global context while adaptively reallocating resolution toward target-relevant regions, resulting in substantially improved counting performance. On the PerSense-D benchmark, AdaCount again achieves SOTA MAE and RMSE, demonstrating its effectiveness across diverse object categories and imaging domains.

\noindent\textbf{PrACo.} We report results on the PrACo benchmark, which comprises the Negative-Label and Mosaic tests designed to assess a model’s ability to correctly interpret textual prompts. As shown in Table~\ref{tab:praco_results}, AdaCount achieves an NMN score comparable to the SAM3 baseline, indicating that the proposed similarity-guided adaptations do not introduce additional false-positive counts for categories absent from the image. At the same time, AdaCount improves PCCN, demonstrating stronger alignment between the target prompt and the predicted counts. On the Mosaic Test, AdaCount consistently improves CntP, CntR, and CntF1 over SAM3, indicating better discrimination between target and distractor categories in multi-class scenes. Interestingly, CountGD++ exhibits a substantially higher NMN score, suggesting weaker prompt awareness despite best counting performance among training-based approaches. This trend is also observed across other trained methods, highlighting that standard counting metrics alone may not adequately capture a model’s ability to distinguish between textual prompts. In contrast, AdaCount preserves the strong prompt-awareness characteristics of SAM3 while consistently improving ZOC performance.

\begin{figure}[t]
    \centering
    \includegraphics[width=1\linewidth]{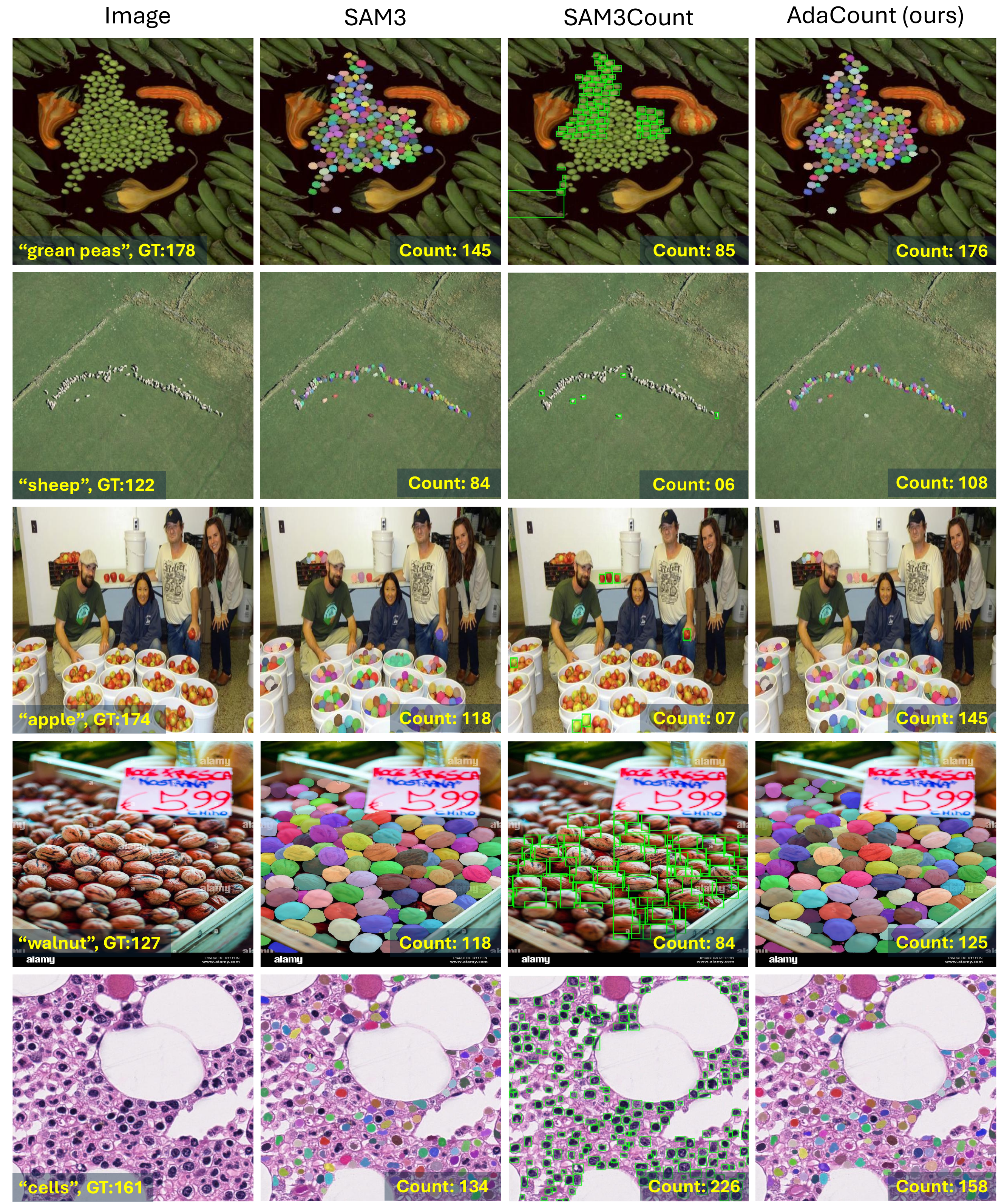}
    \caption{Comparison of SAM3, SAM3Count, and AdaCount. SAM3 frequently undercounts small and crowded objects, while SAM3Count exhibits inconsistent behavior across scenes, including undercounting (row 2-3) and overcounting (last row). In contrast, AdaCount consistently improves counting performance by preserving global context while enhancing target-relevant regions.
}
    \label{fig:qualitative}
\end{figure}

\noindent\textbf{Qualitative Results. } Fig.~\ref{fig:qualitative} presents a qualitative comparison between SAM3, SAM3Count, and AdaCount for the ZOC task. SAM3Count activates its adaptive tiling strategy only when a scene is classified as sufficiently dense. Consequently, challenging scenes that are not identified as dense may receive limited benefit from the refinement stage, leading to substantial undercounting. Furthermore, because tiles are processed independently, contextual information outside the selected region is unavailable during refinement. This limitation is particularly evident in the microscopy example (Fig.~\ref{fig:qualitative}), where aggressive upsampling results in multiple detections within individual cells, leading to overestimated counts. In contrast, AdaCount
preserves global context while increasing representational capacity in target-relevant regions, leading to accurate counts (see Appendix~\ref{appendix} for additional qualitative results.).

\subsection{Ablations}
\label{sec:ablations}

\noindent\textbf{Contribution of Similarity-Guided Adaptations.} Table~\ref{tab:adacount_components} quantifies the contribution of each proposed adaptation. Introducing similarity-guided spatial warping substantially improves performance over the SAM3 baseline, highlighting the effectiveness of reallocating image resolution toward target-relevant regions. Adding feature modulation consistently provides further improvements, demonstrating that the proposed spatial and feature adaptations are complementary. The full AdaCount configuration achieves the best performance, validating the benefit of jointly adapting both the input image and encoder representations.

\noindent\textbf{Sensitivity to Hyperparameters.} We evaluate the sensitivity to key hyperparameters in AdaCount. Table~\ref{tab:gamma_analysis} investigates the similarity sharpening parameter ($\gamma$), which suppresses weak similarity responses while preserving strong responses, thereby increasing the contrast between target-relevant and background regions. The best performance is achieved with ($\gamma=2$). Table~\ref{tab:alpha_analysis} evaluates the feature modulation strength $\alpha$, where performance consistently improves with increasing $\alpha$, reaching its optimum at $\alpha=0.9$. Finally, Table~\ref{tab:gaussian_analysis} examines the effect of Gaussian smoothing, with the best results obtained at $G_\sigma=11$.

\begin{figure}[t]
    \centering
    \includegraphics[width=0.95\linewidth]{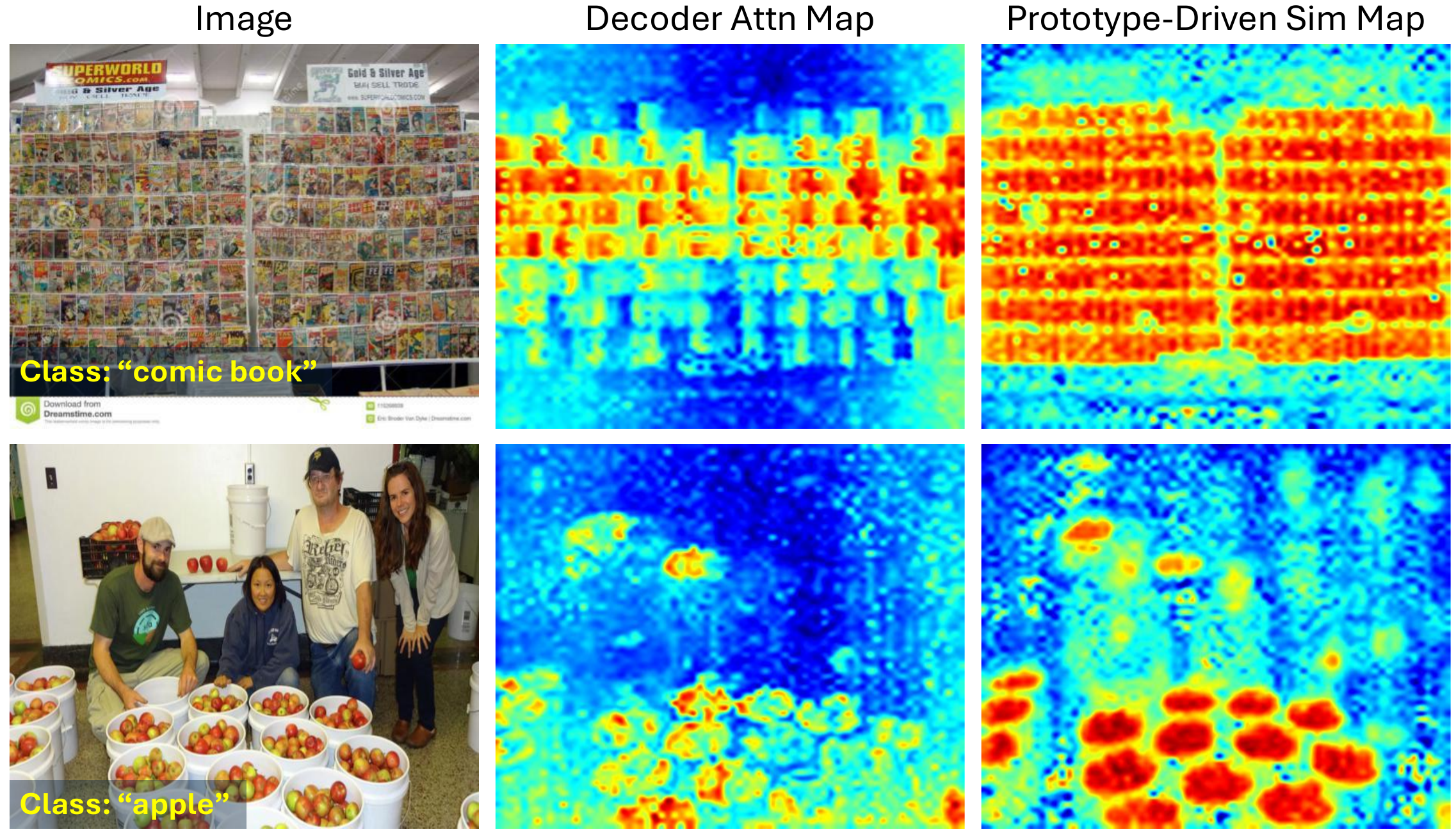}
    \caption{ Attention map produces coarse region-level responses concentrated on dominant semantic structures, whereas prototype-driven similarity map yields localized, instance-aware responses.
}
\vspace{-2mm}
    \label{fig:dvspdm}
\end{figure}

\noindent\textbf{Prototype-Driven Similarity vs Decoder Attention.} We compare the proposed prototype-driven similarity map with a text-conditioned decoder attention map extracted from the last three decoder layers of the first SAM3 pass. The resulting attention map is used to guide both spatial adaptation and feature modulation, while the remainder of the pipeline remains unchanged. As shown in Table~\ref{tab:pds_adaptation}, replacing the proposed similarity map with decoder attention leads to a noticeable performance drop. The qualitative comparison in Fig.~\ref{fig:dvspdm} illustrates the different characteristics of the two signals. Decoder attention produces coarse responses concentrated on a few dominant semantic regions, whereas the similarity map propagates visual evidence from high-confidence target instances to all visually similar instances, producing more localized and instance-aware responses.

\begin{table}[t]
\centering
\footnotesize
\caption{
[Left]: Comparison of prototype-driven similarity (PDS) map and decoder attention (DA) map as guidance signal in AdaCount. [Right]: Runtime efficiency. $N$ denotes number of generated tiles in SAM3Count. Inference cost scales linearly with $N$.
}

\begin{subtable}[t]{0.53\columnwidth}
\centering
\setlength{\tabcolsep}{2pt}
\label{tab:density_adaptation}
\begin{tabular}{lcc}
\toprule
\rowcolor{lightgray!30}
Method & MAE~$\downarrow$ & RMSE~$\downarrow$ \\
\midrule
AdaCount (DA) & 20.23 & 129.62 \\
\rowcolor{lightgreen!50}
AdaCount (PDS) & 18.97 & 126.94 \\
\bottomrule
\end{tabular}
\end{subtable}
\hfill
\begin{subtable}[t]{0.42\columnwidth}
\centering
\setlength{\tabcolsep}{2pt}
\label{tab:inference_cost}
\begin{tabular}{lc}
\toprule
\rowcolor{lightgray!30}
Method & Runtime (ms) \\
\midrule
TFOC & 2100 \\
SAM3 & 300 \\
SAM3Count & $300 \times N$ \\
\rowcolor{lightgreen!50}
AdaCount & 750 \\
\bottomrule
\end{tabular}
\end{subtable}
\vspace{-2mm}

\label{tab:pds_adaptation}
\end{table}


\noindent\textbf{Runtime Efficiency.} Table~\ref{tab:pds_adaptation} reports the average inference time per image for AdaCount and competing methods on an NVIDIA RTX 5000 GPU. AdaCount requires approximately 750 ms per image, compared to 300 ms for vanilla SAM3, reflecting the cost of its two-pass inference pipeline. Despite this modest overhead, AdaCount consistently delivers substantial improvements in counting accuracy. Compared to TFOC (2100 ms), AdaCount is significantly faster. In contrast, the runtime of SAM3Count is scene-dependent and scales with the number of generated tiles, as each tile requires an additional SAM3 evaluation pass. Consequently, inference can become prohibitively expensive in dense scenes; for example, the \emph{``apple''} image in Fig.~\ref{fig:introfig} required approximately 8 minutes to process with SAM3Count. Overall, AdaCount achieves a favorable balance between counting accuracy and inference efficiency.


\noindent\textbf{Limitations.} AdaCount builds upon the SAM3 framework and is therefore inherently bounded by the quality of its initial predictions. In particular, it relies on high-confidence detections from the initial SAM3 inference pass to construct prototype representations. Consequently, if the initial pass fails to detect representative target instances, the subsequent similarity-guided adaptations become ineffective. We observed this failure mode for a few FSC-147 categories, such as \emph{crab cakes} and \emph{finger foods}, where SAM3 failed to produce detections even under relaxed confidence thresholds. Addressing such cases may benefit from more descriptive prompts; however, we use benchmark-provided category names to ensure a fair comparison with existing methods.


\section{Conclusion}
\label{sec:conclusion}

We presented AdaCount, a training-free framework for ZOC that leverages a prototype-driven similarity map to perform similarity-guided spatial warping and feature modulation. By adaptively reallocating representational capacity toward target-relevant regions while preserving global image context, AdaCount consistently improves counting accuracy. Extensive experiments across diverse benchmarks demonstrate that AdaCount consistently outperforms existing training-free ZOC approaches. We hope this work motivates future research on inference-time adaptation strategies for enhancing the counting capabilities of vision foundation models.




%% file: sec/5_Appendix.tex
\clearpage

\appendix

\section{Additional Qualitative Results}
\label{appendix}

\begin{figure*}[t]
    \centering
    \includegraphics[width=1\linewidth]{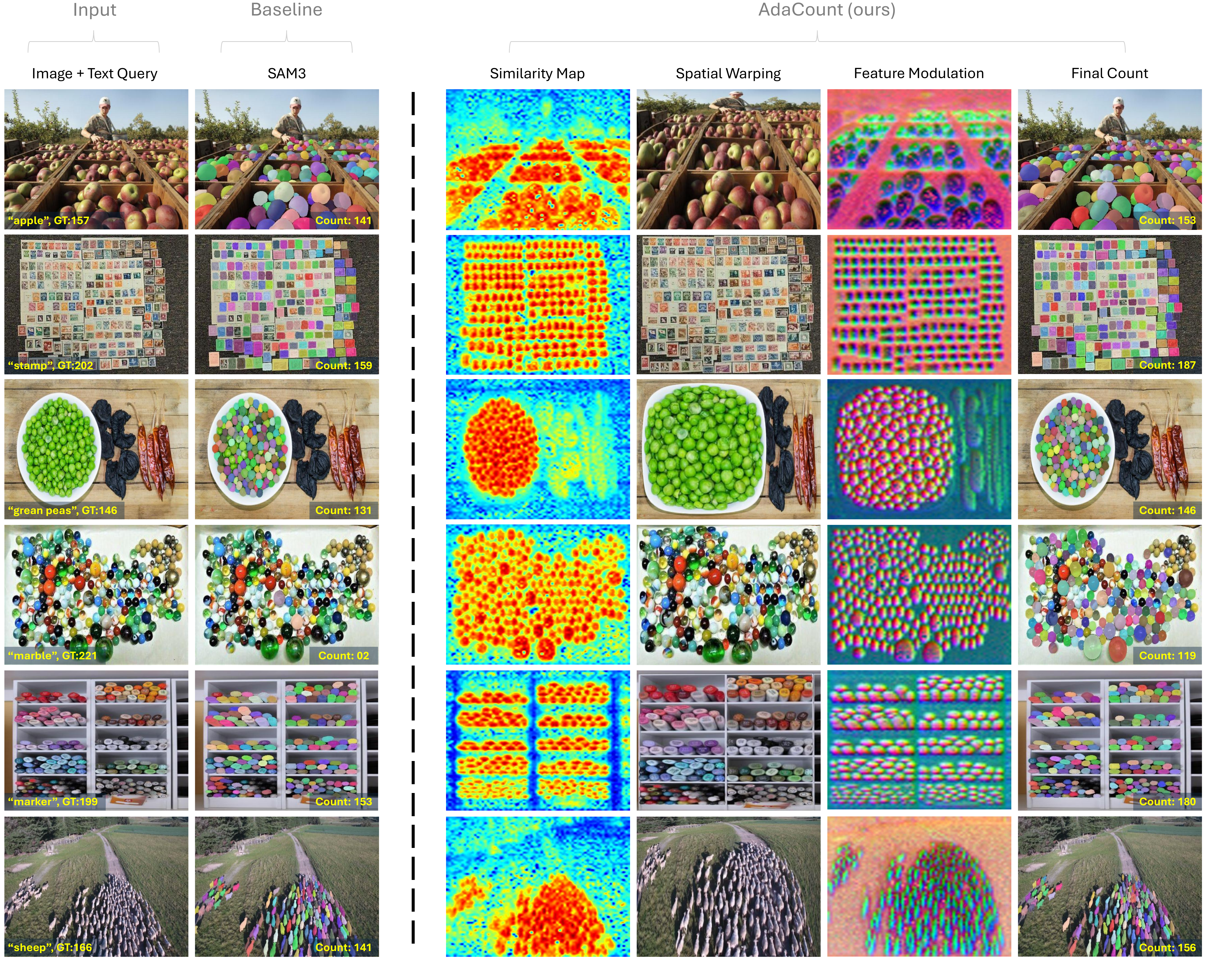}
    \caption{\textbf{Qualitative comparison of SAM3 and AdaCount.} From left to right: \textbf{(1)} input image and text prompt; \textbf{(2)} predictions produced by the vanilla SAM3 baseline; \textbf{(3)} the prototype-driven similarity map estimated from the initial SAM3 inference pass, highlighting target-relevant regions; \textbf{(4)} the similarity-guided spatially warped image, where image resolution is redistributed toward target instances; \textbf{(5)} visualization of the similarity-modulated SAM3 encoder features after feature modulation; and \textbf{(6)} the final AdaCount predictions. Guided by the prototype-driven similarity map, AdaCount performs both spatial warping and feature modulation, producing more discriminative feature representations that ultimately improve counting accuracy compared to the vanilla SAM3 baseline.
}
    \label{fig:qualitative_supp}
\end{figure*}

Fig.~\ref{fig:qualitative_supp} provides additional qualitative comparisons between the vanilla SAM3 baseline and the proposed AdaCount across diverse zero-shot object counting scenarios. In addition to the final counting results, we visualize the intermediate outputs of AdaCount, including the prototype-driven similarity map, the similarity-guided spatially warped image, and the similarity-modulated encoder features. These visualizations illustrate how the estimated similarity map guides both spatial warping and feature modulation, enabling AdaCount to allocate greater representational capacity to target-relevant regions while preserving global image context. As a result, AdaCount consistently improves counting performance relative to the SAM3 baseline.

